\documentclass[runningheads]{llncs}
\usepackage{amsmath,amssymb,amsfonts}
\usepackage{booktabs}
\usepackage{times}

\usepackage{graphicx}
\usepackage{xcolor}
\usepackage{xspace}
\usepackage{url}
\usepackage{multirow}
\usepackage{subcaption}

\newcommand{\varA}[1]{{\operatorname{#1}}}
\newcommand{\ie}{\emph{i.e.,}\xspace}
\newcommand{\eg}{\emph{e.g.,}\xspace}

\newcommand{\modelname}{{EDU-Attention}\xspace}
\newcommand{\rest}{{SemEval-2014 Task-4 Restaurant Review}\xspace}
\newcommand{\laptop}{{SemEval-2015 Task-12 Laptop Review}\xspace}

\begin{document}
\title{Aspect-based Sentiment Analysis through EDU-level Attentions}
%
%
\author{Ting Lin\inst{1} \and
Aixin Sun\inst{1} \and
Yequan Wang\inst{2}}
\authorrunning{Lin et al.}
%
\institute{
School of Computer Science and Engineering, Nanyang Technological University, Singapore\\
\email{ting005@e.ntu.edu.sg}, \email{axsun@ntu.edu.sg}\\ 
\and
Institute of Computing Technology, Chinese Academy of Sciences, China\\
\email{wangyequan@ict.ac.cn}}
\maketitle              
%
%

\begin{abstract}
A sentence may express sentiments on multiple aspects. When these aspects  are associated with different sentiment polarities, a model's accuracy is often adversely affected. We observe that multiple aspects in such hard sentences are mostly expressed through multiple clauses, or formally known as \textit{elementary discourse units} (EDUs), and one EDU tends to express a single aspect with unitary sentiment towards that aspect. In this paper, we propose to consider EDU boundaries in sentence modeling, with attentions at both word and EDU levels. Specifically, we highlight sentiment-bearing words in  EDU through word-level sparse attention. Then at EDU level, we force the model to attend to the right EDU for the right aspect, by using EDU-level sparse attention and orthogonal regularization. Experiments on three benchmark datasets show that our simple EDU-Attention model outperforms state-of-the-art baselines. Because EDU  can be automatically segmented  with high accuracy, our model can be  applied to sentences directly without the need of manual EDU boundary annotation.   
\end{abstract}

\section{Introduction}
\label{sec:intro}

Aspect-based sentiment analysis (ABSA) is challenging because a sentence may express complex sentiments towards multiple aspects. We call these sentences \textit{hard sentences}. For example, ``\textit{Despite the waiter's mediocre \underline{service}, the \underline{food} is tasty and the \underline{bill} is never too large.}" mentions three aspects: service, food, and price, and they are associated with different sentiment polarities. Because an aspect may not always be explicitly expressed through such representative terms, ABSA has been approached by dividing this challenging task into subtasks, \eg to identify aspects in a sentence, and to predict sentiment polarities of the identified aspects. In this paper, we focus on the latter, also known as \textit{aspect category sentiment analysis} (ACSA). In ACSA, the aspects expressed in a sentence are given, and the task is to predict the corresponding sentiment on each given aspect. Note that, the aspect in ACSA is an abstractive category label (\eg price). The name of such an aspect category may not literally appear in a sentence (\eg bill is never large). Our task is different from aspect term-based sentiment analysis (ATSA), where the aspect indicative term in the input sentence are pre-annotated. 



\begin{table}[t]
	\centering
	\small
		\caption{Each of the three 3 EDUs expresses a clear sentiment on one aspect.}
	\label{tab:edu_sentence_example}
	\begin{tabular}{l|c|c}
		\toprule
		Elementary discourse unit (EDU) & Aspect & Polarity\\
		\midrule
		$e_1$. Despite the waiter's mediocre service,~~ & Service & Neutral\\
		$e_2$. the food is tasty,  & Food & Positive \\
		$e_3$. and the bill is never too large. & Price & Positive\\
		\bottomrule
	\end{tabular}

	\vspace{-2ex}
\end{table}
In the example sentence, 
sentiments to the three aspects are expressed in three clauses, or more formally \textit{Elementary Discourse Units} (EDUs), shown in Table~\ref{tab:edu_sentence_example}. EDUs are clause-like grammatical units for discourse parsing in rhetorical structure theory (RST)~\cite{conf/hlt/MannT86}. One EDU carries coherent semantic meaning towards a subtopic~\cite{conf/coling/HuberC20,conf/aaai/JiangFCLZ021}. Thanks to the development of neural models, EDU segmentation can be achieved automatically with high accuracy~\cite{conf/ijcai/LiSJ18}. 

From three benchmark datasets (see Experiments), we observe that, \textit{\textbf{an EDU tends to express at most one aspect}} and \textit{\textbf{unitary sentiment polarity}} towards an aspect. Motivated by this observation, we propose the \textbf{\modelname} model.  Our model learns aspect-specific representation for each EDU independently, as a part of a full sentence representation. Because of  single aspect and unitary sentiment in one EDU, we apply sparse self-attention to select only relevant words to the target aspect in an EDU and ignore irrelevant ones. Considering all EDUs in one sentence, we apply EDU-level sparse self-attention to select the correct EDU(s) for a target aspect. As each EDU only describes a single aspect,  we further apply Orthogonal Regularization on EDU-level attention scores to diversify the attention distributions among all aspects, \ie to ensure that the same EDU is not selected for more than one aspect. 

The \modelname is simple and effective in handling hard sentences in ACSA. Experiments show that our model achieves better accuracy than BERT based models on hard sentences with a much smaller model size and faster inference time.  

\section{Related Work}
\label{sec:related_works}
Predicting sentiment at aspect level is a fine-grained task. 
Attention mechanism~\cite{journals/corr/BahdanauCB14}, as a way of extracting important features from an input sentence, has shown its success in previous studies. A line of work use target aspect as a `query' on terms in an input sentence, to give more weights to aspect relevant terms~\cite{emnlp/TangQL16,emnlp/JiangCXAY19,conf/naacl/XuLSY19,conf/naacl/SunHQ19}. 
There are also works that try to fuse target aspect representation with each term in the sentence before applying attention~\cite{emnlp/WangHZZ16,acl/LiX18,emnlp/HuZZCSCS19,ijcai/LinYL19}. The word-level aspect and term feature fusion makes the input to a model to be more target-specific. 

Syntactic dependency between an aspect and its corresponding opinion expression has also been explored \cite{conf/acl/LiCFMWH20,conf/acl/OhLWPSKK20,conf/naacl/TianCS21}. By utilizing additional syntactic knowledge obtained from external syntax parsers, the relative position in a syntactic tree is used to measure the distance between aspect-related terms and opinion-bearing text span in the sentence. 
These approaches require terms that describe the target aspect explicitly appearing in the sentence, and are pre-annotated. They are not applicable in this work.

Several studies utilize discourse structure/relationship for sentiment analysis. Authors in \cite{ijcnlp/ZirnNSS11,acl/LazaridouTS13} explore discourse relationship between two adjacent EDUs for predicting sentiment polarity. 
Hand-crafted rules are used to segment text into sentiment expression units (SEUs); a SEU contains either a sentiment, or an aspect, or both~\cite{emnlp/ZhangS18}. In \cite{icwe/HoogervorstEJHS16}, a full discourse parse tree is utilized to find precise context for a given aspect term. 
There are also neural network approaches that utilize EDU segments for document-level sentiment prediction~\cite{tacl/AngelidisL18,ijcai/WangLLKZSZ18,tkde/li2020neural}. The overall sentiment polarity becomes an aggregation of sentiment distribution of EDUs in the document. 
Different from these models, we do not consider relationships between adjacent EDUs or their locations in a discourse parse tree. Instead, we model each EDU independently and apply word-level sparse attention to give more weights to relevant terms. By assuming one EDU expresses at most one aspect, we use regularisation at EDU-level attention to avoid the same EDU being selected for multiple aspects. 

\section{The Proposed Model: EDU-Attention}
\label{sec:model}



We follow the definition of \textit{aspect category sentiment analysis} (ACSA) in previous studies~\cite{www/WangSHZ19,emnlp/LiYZP20,emnlp/JiangCXAY19}. There are  $k$ predefined aspect categories $A = \{ a^1, \dots, a^k \}$,  and a list of predefined sentiment polarities $P = \{negative, neutral, positive\}$. Given a sentence $s$ and the $m$ aspect label(s) expressed in the sentence $A_{s} = \{a^{1}, \dots, a^{m}\}, A_s \subseteq A$, we aim to predict the sentiment polarity associated with each aspect label, \ie all pairs of $(a^{m}, p_{o})$, for  $a^{m} \in A_s, p_{o} \in P$. 

\subsection{Model Overview}

\begin{figure}[t]
    \centering
	\begin{subfigure}[b]{\columnwidth}
		\includegraphics[width=\linewidth]{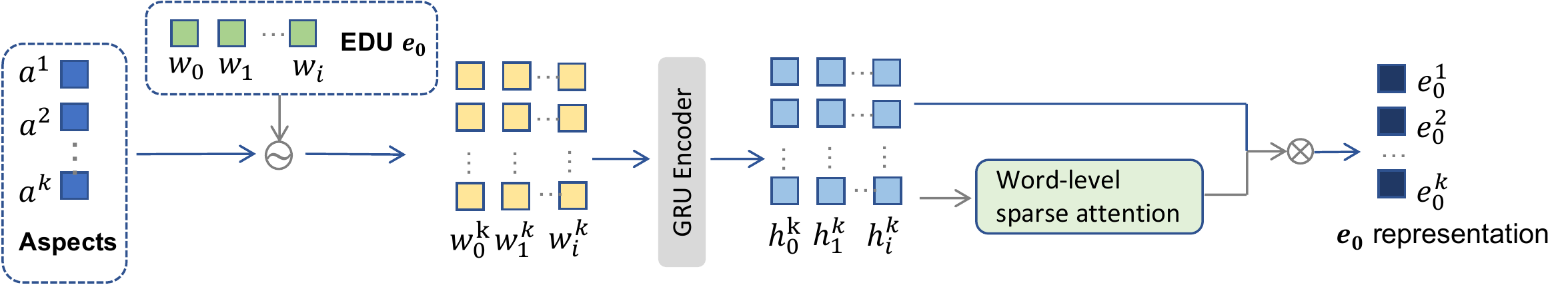}
		\caption{A $k$-aspect-specific EDU representation is learned for the given $k$ aspects. Word-level sparse attention is applied to give more weights to sentiment-bearing words.}
		\label{sfig:model_edu}
	\end{subfigure}
	\vspace{1ex}
	\begin{subfigure}[b]{\columnwidth}
		\includegraphics[width=\linewidth]{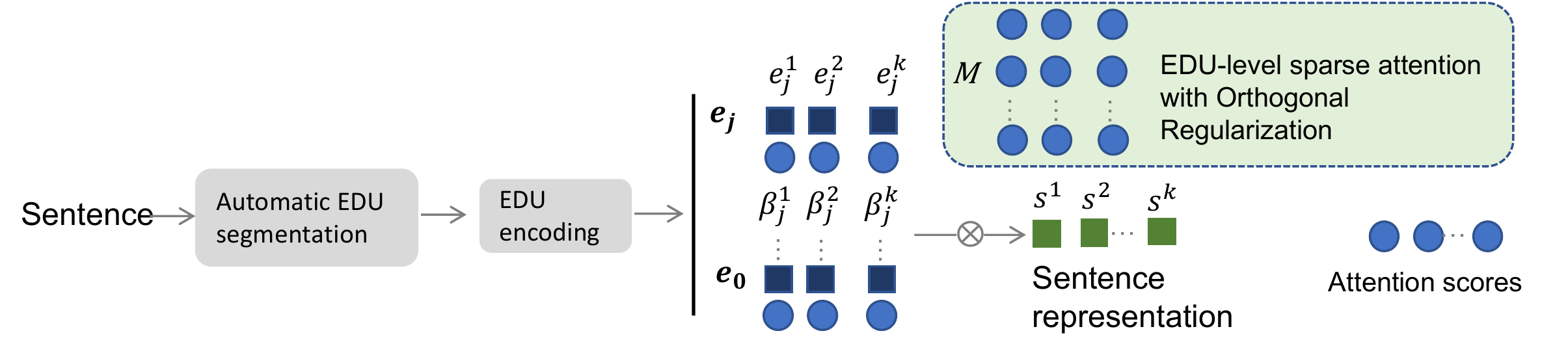}
		\caption{Aspect-specific sentence representation is a linear sum of EDU representations and their respective attention weights, on each aspect. Sentiment is predicted on the corresponding $s^k$; detecting the existence of the aspect is an additional learning objective.}
		\label{sfig:model_sentence}
	\end{subfigure}
	\vspace{-2ex}
	\caption{The model architecture of \modelname.}
	\label{fig:model_architectures}
\vspace{-2ex}
\end{figure}

\modelname is a simple model that takes in an EDU-segmented sentence as input. EDU segmentation of a sentence can be achieved with high accuracy by using off-the-shelf tools.\footnote{\scriptsize In our implementation, we use the pretrained \textsc{SegBot} tool  \url{http://138.197.118.157:8000/segbot/} released by its authors~\cite{conf/ijcai/LiSJ18}. If an EDU returned by \textsc{SegBot} contains conjunction words (\ie `but',`and', `although', and `or'), we further split this EDU by using regular expression.} After \textit{automatic EDU segmentation}, an input sentence is denoted by its EDUs $\{\mathbf e_0, \dots, \mathbf e_j\}$, and an EDU is a sequence of words $\mathbf e_j=\{w_0, \dots, w_i\}$.

Fig.~\ref{sfig:model_edu} shows the EDU-encoder in our model. It learns an aspect-specific EDU representation for a given EDU. Specifically, an EDU is represented as a $k$-dimensional feature, one for each of the $k$ aspects. The encoder applies sparse self-attention to words within the EDU, to give sentiment-bearing words more weight for that particular aspect. Shown in Fig.~\ref{sfig:model_sentence}, with all aspect-specific EDUs representations learned in a sentence, we apply EDU-level sparse self-attention for locating the right EDU(s) for the right aspect. We further apply orthogonal regularization to force different aspects to focus on different EDUs. Finally,  sentence representation is obtained as a linear sum of the EDUs representations with respective attention scores. We use aspect category prediction as an auxiliary learning objective, in addition to sentiment label prediction.
\subsection{EDU Representation}
The EDU-encoder learns aspect-specific EDU-representation for an EDU in three steps: word-aspect feature fusion, EDU encoder, and word-level sparse attention, see Fig.~\ref{sfig:model_edu}. 

\paragraph{Word-aspect feature fusion.} To promote interaction between aspect and words, we fuse word embedding $w_{i}$ with aspect embedding $a^{k}$ to derive aspect-specific word feature $w_{i}^{k}$, shown in Eq.~\ref{eq:edu_word_fusion}. Here, $W_{e}$ and $W_{k}$ shared for all aspects are learnable parameters. Aspect embedding $a^{k}$ can be initialized by using embedding of a matching word for the aspect (\eg `food', `service'), or be initialized randomly, if there is no representative word for the aspect (\eg `anecdotes/miscellaneous'). 
\begin{equation}
	w_{i}^{k} =\tanh\left(w_{i}W_{e} + a^{k}W_{k}\right)
\label{eq:edu_word_fusion}
\end{equation}	
\paragraph{EDU encoder.} With the fused aspect-word features, we perform EDU encoding by using a bidirectional $\varA{GRU}$ shown in Eq.~\ref{eq:edu_encoding}. The bidirectional $\varA{GRU}$ learns contextual information within an EDU, for each aspect. Specifically, the encoder takes in an aspect-word feature matrix of the EDU $e_j$ for aspect $k$, and stacks hidden output from every time step of $\varA{GRU}$ into $H_{j}^{k}$ for further processing. 
\begin{equation}
	H_{j}^{k}=\mathrm{\varA{GRU}}([w_{0}^{k}, w_{1}^{k}, \dots, w_{i}^{k}])
	\label{eq:edu_encoding}
\end{equation}
\paragraph{Word-level sparse attention.} Ideally, a classifier only needs to extract aspect-specific opinion-bearing words for predicting the sentiment. The remaining words can be ignored. We apply the attention mechanism on $H_{j}^{k} $ to highlight such important words. 

A straightforward solution is to apply self-attention with $\mathrm{softmax}$ normalization~\cite{journals/corr/BahdanauCB14}. The resulting probability is distributed to all words in an EDU, \ie the attention score after $\mathrm{softmax}$ is not equal to $0$ for every word $w_{i}\in \mathbf e_j$. This does not well serve our purpose of ignoring irrelevant words in the EDU. Hence, we adopt $\mathrm{sparsemax}$ function~\cite{icml/MartinsA16}, which returns the euclidean projection of the $k$ element input vector $z$ onto the $(k-1)$-dimensional simplex $\bigtriangleup^{k-1}$ defined as $\{p\in \mathbb{R}^k| 1^Tp=1, p\geq0 \}$~\cite{icml/MartinsA16}. The projection is likely to hit the boundary of the simplex, in which case the $\mathrm{sparsemax}(z)$ becomes sparse. $\mathrm{sparsemax}$ produces sparse distribution while retaining the important properties of $\mathrm{softmax}$. Eq.~\ref{eq:sparse_max} shows the sparse attention computation. 
\begin{equation}
	\mathrm{sparsemax}(z) = \underset{p\in \bigtriangleup^{k-1}}{\mathrm{argmin}} ||p-z||^2
	\label{eq:sparse_max}
\end{equation}
By using  feature matrix $H_{j}^{k}$ obtained earlier, we apply $\mathrm{sparsemax}$ to compute a weight vector for all words in EDU $\mathbf e_j$, for aspect $a^{k}$. The sparse attention computation, where $\alpha_{i}^{k}$ is the attention score of word $w_i$ in EDU $\mathbf e_j$ towards aspect $a^k$ is calculated by:
\begin{equation}
	\alpha^{k}_{0}, \dots, \alpha^{k}_{i} = \mathrm{sparsemax}(H^{k}_{j}W_e),
	\label{eq:sparse_max_edu}
\end{equation}
where $W_e$ is a learnable parameter.

Then, for EDU $\mathbf e_j$, we derive the $k$ number of EDU representations $[e_j^1,\dots,e_j^k]$, one for each aspect. For aspect $a^k$, $e_j^k$ is the weighted sum of the sparse attention scores $\alpha_{i}^{k}$'s and $h_i^k$'s for all the words in $\mathbf e_j$. We also add position embedding~\cite{conf/naacl/DevlinCLT19} of EDU $\mathbf e_j$, denoted by $\mathrm{Emb}_p(\mathbf e_j)$ , for its relative position in the sentence, as shown in Eq.~\ref{eq:edu_rep}.
\begin{equation}
		e^k_{j} = \mathrm{Emb}_p(\mathbf e_j) + \sum_{w_i \in \mathbf e_j} \alpha_{i}^{k}  h^k_{i}
	\label{eq:edu_rep}
\end{equation}

\subsection{Sentence Representation and Learning Objective}

So far, for an EDU $\mathbf e_j$, we obtain its aspect-specific representation $[e_j^1,\dots,e_j^k]$. As we observe that one EDU tends to express sentiment on one aspect, we now try to identify the right aspect for each EDU in a sentence. 
\paragraph{EDU-level sparse attention.} For each aspect $a^k$, we apply sparse attention on the corresponding aspect-specific EDU-representations in the sentence $[e_0^k, \dots, e_j^k]$, for choosing the right EDU for this aspect. The sparse attention shares the similar process as word-level sparse attention, or formally:
\begin{equation}
	\beta_{0}^k, \dots, \beta_{j}^k = \mathrm{sparsemax}([e_0^k, \dots, e_j^k]W_s),
	\label{eq:edu_attention}
\end{equation} 
where $W_s$ is a learnable parameter. Then, the aspect $a^k$ specific sentence representation $s^k$ is a linear combination of its aspect-specific EDU representations. 
\begin{equation}
		s^k = \sum_{\mathbf e_j\in s} \beta_{j}^k  e^k_{j}
	\label{eq:sentence_rep}
\end{equation}
Each $s^k$ is used to predict sentiment label, and also to predict the corresponding aspect as an additional objective. 

\paragraph{The aspect-level orthogonal regularization.} As stated earlier, an EDU tends to express a single aspect and a unitary sentiment. In a complex sentence, opinions for different aspects reside in different EDUs. The sparse attention computed in Eq.~\ref{eq:edu_attention} is for one specific aspect, and attention scores for different aspects are computed independently.

To constraint that one EDU should be attended to a single aspect, we put the EDU-level attention scores $\beta_j^k$'s computed for the $k$ aspects over the $j$ EDUs in a sentence, into a $j \times k$ attention matrix $M$. Then we apply orthogonal regularization to force the dot product of attention vectors of each aspect to be orthogonal, as shown in Eq.~\ref{eq:orthoganal_regularization}. $I$ is an identity matrix.
\begin{equation}
	\begin{aligned} 
		R_{orth} = \left\|M^{T}M - I \right\|
	\end{aligned}
	\label{eq:orthoganal_regularization}
\end{equation}
\paragraph{Learning objectives.} The key objective is to predict sentiment polarities for the given aspects mentioned in a given sentence.  In \modelname, we also use aspect prediction as an additional learning objective in addition to the sentiment labels. Specifically, a binary prediction on each aspect existence in the input sentence is added into the loss function of our model. We use cross-entropy loss $J(\theta)$ for sentiment labels prediction, and binary cross-entropy loss $U(\theta)$ for aspect prediction. The aspect-level orthogonal regularization $R(\theta)_{orth}$ is also a part of our learning objectives. The full loss function of our model is as follows:
\begin{equation}
	 L(\theta) = \lambda_1 J(\theta) + \lambda_2 U(\theta) + \lambda_3 R(\theta)_{orth},
	\label{eq:objectives}
\end{equation}
where $\lambda_1,\lambda_2,\lambda_3$ are the scaling parameters set for each loss.  The collection of model parameters is $\theta$.

\section{Experiments}
\label{sec:exp}
We evaluate the proposed \modelname on three benchmark datasets, with a focus on hard sentences.

%

\begin{table}[t]
	\centering
	\small	
		\caption[Dataset statistics]{Statistics of  datasets, with \#sentences expressing single and multiple aspects. We remove sentence with conflict polarities.}
	\label{tbl-datasets}
	\begin{tabular}{l@{\hspace{2mm}}|l@{\hspace{2mm}}l@{\hspace{2mm}}|c@{\hspace{2mm}}|l@{\hspace{2mm}}l@{\hspace{2mm}}|c@{\hspace{2mm}}|l@{\hspace{2mm}}l@{\hspace{2mm}}l@{\hspace{2mm}}}
		\hline
		  \multirow{2}{*}{Datasets} & \multicolumn{2}{c|}{{Rest14}}& {Rest14-Hard} & \multicolumn{2}{c|}{{Laptop15}}  & {Laptop15-Hard} & \multicolumn{3}{c}{{MAMS-ACSA}}\\
		\cmidrule{2-10}
		  & Train & Test & Test & Train & Test & Test & Train & Val & Test \\ 
		\toprule
		Single & 2,345 & 595 & - & 1,174 & 539 & - & - & - & - \\
		Multiple & 539 & 172 & 25 & 209 & 93 & 20 & 2,839 & 710 & 400 \\
		\midrule
		Negative & 841 & 222 & 20 & 616 & 258 & 14 & 1,883 & 460 & 263 \\ 
		Neutral & 501 & 94 & 12 & 58 & 44 & 8 & 2,776 & 689 & 393 \\ 
		Positive & 2,174 & 657 & 21 & 860 & 424 & 19 & 1,742 & 428 & 263 \\
	  \bottomrule
	\end{tabular}
	\vspace{-2ex}
\end{table}
\subsection{Datasets and Baselines}
\label{ssec:datasetbaseline}

Table~\ref{tbl-datasets} summarizes the three datasets in our experiments. Following previous studies~\cite{emnlp/TangQL16,emnlp/WangHZZ16}, we remove samples with conflict polarities.  \textbf{Rest14} is from \rest~\cite{semeval/PontikiGPPAM14}. 
\textbf{Rest14-Hard} is a collection of hard sentences sampled from Rest14 test set~\cite{acl/LiX18}. Each sentence in Rest14-Hard contains at least two aspects, and the aspects have different sentiment polarities. \textbf{Laptop15} is from \laptop~\cite{semeval/PontikiGPPAM14}. 
To be consistent with other datasets, we keep aspects only and ignore attributes. Accordingly, we update sentiment labels of aspects following the original annotation guideline~\cite{semeval/PontikiGPPAM14}.\footnote{\scriptsize \url{https://alt.qcri.org/semeval2014/task4}} We choose to keep the aspects that contain at least one sentence in every sentiment class (positive, neutral, and negative). In total, there are 22 aspects.\footnote{\scriptsize We will release the re-processed Laptop15 dataset.} \textbf{Laptop15-Hard} is a collection of hard sentences sampled from Laptop15 test set, following the same sampling strategy as Rest14-Hard. \textbf{MAMS-ACSA} is a restaurant review dataset~\cite{emnlp/JiangCXAY19}. 
\textit{All sentences in MAMS-ACSA are hard sentences}; 
Note that,  MAMS-ACSA's \textit{annotation scheme is different} from that of other datasets. A sentence is annotated with an aspect if the sentence mentions a matching keyword. 
The differences in annotation scheme results in majority of the sentiment labels being `neutral' in MAMS-ACSA, making the dataset challenging.

We evaluate the following baselines. 
\textbf{ATAE-LSTM}~\cite{emnlp/WangHZZ16}, is a strong baseline where aspect embeddings are concatenated with word vectors. 
\textbf{MemoryNet}~\cite{emnlp/TangQL16}, employs two LSTMs and an interactive attention mechanism to learn representations of  sentence and aspect. 
 \textbf{HAN}~\cite{ijcai/WangLLKZSZ18} is a hierarchical attention network built on word, clause (EDU), and sentence for aspect-specific sentence representation.
 \textbf{GCAE}~\cite{acl/LiX18} uses CNNs to extract features and then employs two Gated Tanh-Relu units to selectively output the sentiment information flow towards the aspect, for predicting sentiment labels.
 \textbf{ATAE-CAN-2$\mathbf{R_o}$}~\cite{emnlp/HuZZCSCS19} uses aspect detection as an auxiliary task. 
 \textbf{AS-Capsule}~\cite{www/WangSHZ19} is a capsule alike network. Each capsule encloses a set of computations for one aspect. 
 \textbf{CapsNet}~\cite{emnlp/JiangCXAY19} is a capsule-network based model. It learns the association between aspect and context.
 \textbf{AC-MIMLLN}~\cite{emnlp/LiYZP20} is a multi-instance learning model. It has two separate encoders to learning aspect- and sentiment- representations. 

These baselines use different approaches of learning aspect-specific sentence representations. ATAE-LSTM, ATAE-CAN, GCAE, HAN fuse aspect features with term features, then apply attention on the fused features. MemoryNet and CapsNet use carefully designed attention mechanisms on term features only. In addition, AC-MIMLLN, ATAE-CAN, and AS-Capsule use aspect category prediction as an auxiliary task for learning the interaction between aspect representation and sentiment representation. 

We also evaluated Bert-based models. The simple \textbf{Bert-baseline} and its `distillation' versions (DistillBERT \cite{journals/corr/abs-1910-01108}, TinyBERT\cite{conf/emnlp/JiaoYSJCL0L20})  encode an aspect-specific sentence representation with this input format: [CLS] words in sentence [SEP] aspect category [SEP] \cite{conf/naacl/XuLSY19}.  The sentiment prediction is done by $\mathrm{softmax}$ with a linear layer. In the Bert variant of our model, we replace the EDU representation learning (\ie Fig.~\ref{sfig:model_edu}) by Bert encoding with: [CLS] words in  EDU [SEP] aspect category [SEP]. For each EDU, we enumerate all aspect categories to obtain its aspect-specific representation.  \textbf{CapsNet-Bert} is a variation of CapsNet; \textbf{BERT-pair-QA-B} \cite{conf/naacl/SunHQ19} constructs an auxiliary sentence for each aspect and transform the task into a sentence-pair classification. For CapsNet-Bert and BERT-pair-QA-B, we use authors' implementation. 

\begin{table*}[t]
 \small
	\centering
		\caption{Accuracy and Macro-F1 of all models on all datasets. We reproduce results of baselines by using authors' implementation except two models. ATAE-CAN-2$R_o$ is by our own implementation following authors' paper. Results of AC-MIMLLN model are reported in its original paper. We run a model 5 times with random seeds and report the average. The best performance are in boldface and second best underlined, among non-Bert and Bert models respectively.}
	\label{lbl-table-results}
	\begin{tabular}{l@{\hspace{1mm}}|c@{\hspace{2mm}}|c@{\hspace{2mm}}|c@{\hspace{2mm}}|c@{\hspace{2mm}}|c@{\hspace{2mm}}|l@{\hspace{2mm}}|c@{\hspace{2mm}}|c@{\hspace{2mm}}|c@{\hspace{2mm}}|c@{\hspace{2mm}}}
		\toprule
		\multicolumn{1}{l|}{\multirow{2}{*}{Model}} & \multicolumn{2}{c|}{Rest14} & \multicolumn{2}{c|}{Rest14-Hard} &  \multicolumn{2}{c|}{Laptop15}&\multicolumn{2}{c|}{Laptop15-Hard}&\multicolumn{2}{c}{MAMS-ACSA}\\
		\cmidrule{2-11}
		\multicolumn{1}{c|}{} & {Acc.} &{$F_{1}$} & {Acc.} &{$F_{1}$}& {Acc.} &{$F_{1}$} & {Acc.} &{$F_{1}$}& {Acc.} &{$F_{1}$}\\
		\midrule
		MemoryNet   	& 81.29 & 70.79& 54.72&46.65& 71.80 & 52.03 &36.59&26.09& 64.04& 62.57 \\					
		AC-MIMLLN* 		& 81.60 & -     & \underline{65.28}& -&- & - &-&-& \underline{76.64}& - \\
		HAN   			& 81.74 & 71.49 &58.49&49.67&73.21 & 53.64 &48.78&41.05& 73.64 & 72.42  \\	
		AS-Capsule 		& 82.03 &71.55 &59.24&51.99& 76.31 & 55.65 &48. 78&41.05& 75.44 & {74.37} \\ 	
		GCAE   			& 82.32 & {72.08}&56.13&51.07&76.08 & 54.00 &\underline{55.29}&42.78& 70.59 & 69.01 \\		
		ATAE-CAN-2$ R_o $ &82.43&71.18&64.62&\underline{53.74}&76.24& 53.00&51.22&\underline{43.66}& 76.42& \underline{75.23}\\
		ATAE-LSTM   	& \underline{83.10} & \underline{73.32} &59.91& 53.08& \underline{76.86} & \textbf{56.88}&46.34&38.35& 75.02 & 73.93\\						
		CapsNet  		& \underline{83.10} & 72.58 &53.78&44.50&75.48&52.33&48.78 &33.65&72.92&71.86\\				 		
		\midrule
		\textbf{Ours} 			& \textbf{83.97}& \textbf{73.96}&\textbf{70.28}&\textbf{65.59}&\textbf{77.83} &\underline{56.39}  &\textbf{56.10}&\textbf{46.35}& \textbf{77.14} & \textbf{76.00}\\	
		Ours w/o reg.   & 82.88 & 72.69 &68.68&60.79& 76.37& 55.05&51.22&41.49& 75.65 	& 74.50  \\	
		Ours w/o aux.   & 82.99 & 73.02 &58.51& 54.49&  77.05	& 55.23 	&53.66&45.08& 75.03 	& 73.91	  \\
		\midrule \midrule
		DistilBERT &65.57&38.36&41.51&26.72&59.09&32.39&53.66&33.73&59.16&47.17\\			
		TinyBERT &67.52&26.87&39.62&18.92&57.44&32.58&56.10&38.02&59.49&47.45\\					
		Bert-baseline&87.82&80.07&66.98&62.83& 83.47 	& 63.94	 	&58.40&24.58& 78.86	&78.06\\
		CapsNet-Bert &87.80&	79.81  &50.94& 38.66& 84.07	& 57.25 	&48.78&33.65& 77.42	& 76.65   \\
		BERT-pair-QA-B &87.25&78.09&52.83&46.58&83.88&\textbf{69.77}&46.34&39.19&79.35&78.89\\ 
		\textbf{Ours-Bert}   	&\textbf{87.94}&	\textbf{80.74}   &\textbf{72.95}& \textbf{70.71}& \textbf{84.85}	& {65.07}	 &\textbf{58.54}&\textbf{41.56}& \textbf{79.64}&	\textbf{79.02}\\
		\bottomrule
	\end{tabular}
	\vspace{-2ex}
\end{table*}
 
\subsection{Implementation and Parameter Setting}
All models are implemented by using Pytorch\footnote{\scriptsize\url{https://pytorch.org/}} with CUDA 11.1 on RTX3090 GPU in Windows OS. Models' parameters are optimized by using Adam. For non-Bert models, we set a learning rate of $1e-3$ for model parameters and $1e-4$ for word embedding adjustment. The word embeddings are initialized by Glove~\cite{emnlp/PenningtonSM14} with 300  dimensions. We set the mini-batch size to 32 and evaluate every 16 mini-batches, and use a dropout rate of 0.5 during model training. We use `bert-base-uncased'~\footnote{\scriptsize{ \url{https://huggingface.co/bert-base-uncased}}} for fine-tuning models use BERT, `distilbert-base-uncased'~\footnote{\scriptsize{\url{https://huggingface.co/distilbert-base-uncased}}} for DistillBERT model, and `TinyBERT\_General\_6L\_768D'~\footnote{\scriptsize{\url{https://huggingface.co/huawei-noah/TinyBERT_General_6L_768D}}} for TinyBERT model. For fine-tuning, we keep dropout probability at $ 0.1 $, learning rate at $3 e-5$. We set the scaling parameter to $\lambda_1,\lambda_2$ to 1, and $\lambda_3$ to 0.1 for \modelname (see Eq.~\ref{eq:objectives}). In the Bert variation of \modelname, we set $\lambda_1,\lambda_2,\lambda_3$ to 0.5, 0.4, and 0.1 respectively for the best performance. Models' parameters are tuned on validation set. The MAMS-ACSA dataset comes with a validation set. For Rest14 and Laptop15 datasets, we randomly sample 20\% of training data as validation set.  We run the models for 5 times with random seed initialization, and report the average metric on test sets. 
\subsection{Comparison with Baselines}

\paragraph{Non-Bert Models.} Reported in Table~\ref{lbl-table-results}, among non-Bert models, our \modelname performs the best on almost all metrics, except Macro-F1 on Laptop15, which is the second best. Large improvements are achieved on Rest14-Hard and Laptop15-Hard. 

The Rest14 dataset contains 19\% multi-aspect sentences (see Table~\ref{tbl-datasets}). 
Models that use aspect and term fused features (\eg ATAE-CAN, ATAE-LSTM, GCAE) generally perform better than others. In particular, ATAE-LSTM performs the second-best among non-bert  models by both accuracy and Macro-F1. 
We produce results on Rest14-Hard by using the models trained on Rest14 training dataset. Our \modelname outperforms all baselines by a large margin on both metrics, showing the effectiveness of modeling EDU contextual boundaries in handling hard sentences. 

The Laptop15 dataset contains 15\% multi-aspect sentences. All models are less affected by noise introduced by sentiment terms of non-target aspects, compared to other datasets. On hard sentences, our model shows clear advantage on Laptop15-Hard.

All sentences in  MAMS-ACSA  are hard sentences.  A different annotation scheme is adopted in MAMS-ACSA as described in the Dataset section. The annotation based on appearance of surface terms, makes the dataset challenging with many neutral labels. Our model is the best performer, demonstrating its effectiveness in handling hard sentences. 

\paragraph{Bert Models.} As expected, Bert-baseline outperforms all non-Bert models and brings in big improvements on Rest14 and Laptop15 datasets. The improvement over \modelname on the MAMS dataset, however, is relatively small. On Rest14-Hard, Bert-baseline performs poorer than \modelname by about 3 points for both Accuracy and Macro-F1. Recall that MAMS, Rest14-Hard, and Laptop15-Hard datasets only contain hard sentences; the comparable performance between \modelname and Bert-baseline suggests that modeling EDU contextual boundary is beneficial to aspect category sentiment analysis. By using Bert for EDU representation learning, our \modelname-Bert model outperforms all models. In particular, on Rest14-Hard, our \modelname-Bert outperforms Bert-baseline by 6 to 7 points on both Accuracy and Macro-F1. We also compare the `distillation' version of BERT (DistillBERT, TinyBERT) with the same input format and model training strategy, the performance of both models does not come close to the rest of the baselines.

\subsection{Ablation Study}
The orthogonal regularization in our model makes the distribution of EDU-level attention scores diverse by aspects.
We also predict the existence of an aspect (\ie $U(\theta)$ in Eq.~\ref{eq:objectives}) as an additional learning objective to enable the model to concentrate more on the aspect relevant EDU(s). 

We conduct ablation study to analyze effectiveness of the orthogonal regularization (reg.) and the auxiliary aspect prediction (aux.) in \modelname. Reported in Table~\ref{lbl-table-results}, removing either orthogonal regularization or  auxiliary aspect prediction leads to performance drop on all datasets and on measures. The amount of performance drop is comparable on Rest14, Laptop15, and MAMS-ACSA datasets. Large drops are observed on Rest14-Hard. On hard sentences, our model relies on orthogonal regularization to spread out the EDU-level attention scores. The additional objective function guides the model to recognize the aspect expressed in an EDU. The amount of drop on the MAMS-ACSA dataset is slightly larger than that on Rest14 and Laptop15. On the one hand, all sentences in MAMS-ACSA are hard sentences. On the other hand, its annotation scheme is different from the other datasets. The model might have captured the association between term appearance and aspects, and the model performance is heavily affected by the large number of neutral labels (see Table~\ref{tbl-datasets}).
\subsection{Analysis of Sparse Attention}
\begin{figure}[t]
	\centering
	\includegraphics[width=.85\columnwidth]{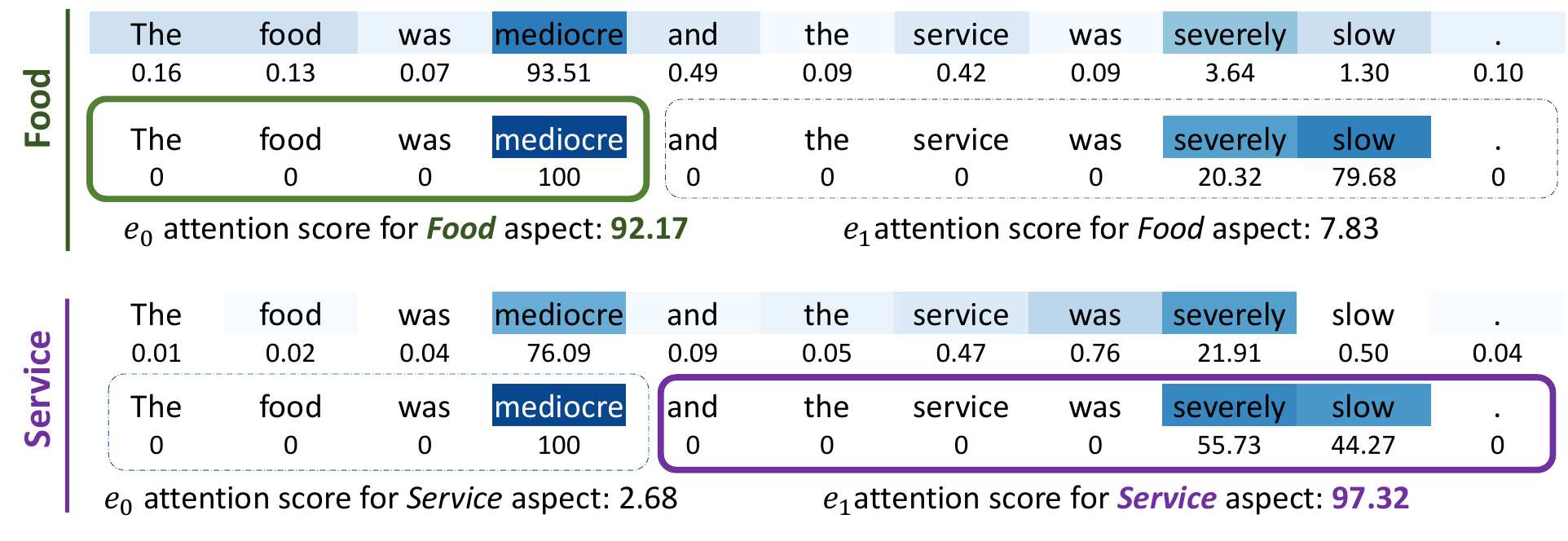}
		\caption{Attention scores from aspect-specific representations of ATAE-LSTM and \modelname, on two aspects food and service. Outputs from ATAE-LSTM is shown in one sentence for each aspect. EDU-Attention has both word- and EDU-level attentions, shown with EDU boundaries.}
		\label{fig:attention-compare}
		\vspace{-2ex}
\end{figure}

As a case study, we compare attention scores computed by ATAE-LSTM \cite{emnlp/WangHZZ16}, and the attention scores of our model at both word- and EDU- levels. Fig.~\ref{fig:attention-compare} shows an example sentence from Rest14. The sentence expresses sentiments towards two aspects: food and service, with indicative words. ATAE-LSTM computes two aspect-specific representations, one for each aspect. In both representations, the model highlights `mediocre' and `severely' with larger scores. Nevertheless, the score of `mediocre' dominates in both representations. Even though ATAE-LSTM also correctly highlights `severely' for aspect `service', the larger score of `mediocre' would interfere with the correct prediction of sentiment label for service. 

In contrast, the word-level attention scores computed by \modelname is confined within an EDU contextual boundary, and normalized within the EDU. 
Opinion words `mediocre' and `severely slow' are given large attention scores. In fact, all the rest non-opinion words are assigned 0 scores, thanks to sparse attention (see Eq.~\ref{eq:sparse_max}). The association between the opinion words and their corresponding aspects is through the EDU-level sparse attention. Recall that, for each EDU, we learn an aspect-specific representation. As shown in Fig.~\ref{fig:attention-compare}, for the first clause (\ie EDU $\mathbf e_0$), its attention score for  `food' aspect is 92.17, compared to 2.68 for  `service' aspect. Note that, EDU-level attention scores are normalized across all EDUs in the same sentence on each aspect. Similarly, the second clause, EDU $\mathbf e_1$, receives high score 97.32 for `service',  and a very small score 7.83 for `food' aspect. In short, sparse attentions at both EDU- and word- levels guide our model to correctly identify aspect-relevant EDU in a sentence, and opinion words in an EDU. 

\subsection{Model Size and Inference Time}
\begin{table}[t]
	\centering
	\small
	\caption{Model size, and inference time (in second) on  MAMS test datset.}
	\label{lbl-model-size-inference-timing}
	\begin{tabular}{l@{\hspace{1mm}}|l@{\hspace{1.5mm}}|l@{\hspace{1mm}}|l@{\hspace{1.5mm}}|l@{\hspace{1mm}}|l@{\hspace{0.5mm}}}
		\toprule
		Model & \#Params & Inference(s) &Model & \#Params & Inference(s)\\
		\midrule 
		MemoryNet & 2.4M& 0.0673 & Distill-Bert &66.3M &0.3679  \\
		GCAE & 2.8M & 0.0775 & Tiny-Bert & 66.9M & 0.3736 \\
		ATAE-LSTM & 3.8M & 0.1298  & Bert-baseline & 109.5M & 0.9065\\
		ATAE-CAN-2$R_0$ &3.8M & 0.3702 & BERT-pair-QA-B & 109.5M &17.496\\
		HAN & 6.0M & 0.2195& CapsNet-Bert &111.9M & 0.9122 \\
		CapsNet &6.0M& 0.9072 & Ours-Bert & 110.7M& 3.5734\\
		AS-Capsule & 10.0M & 0.4215 & \modelname (\textbf{Ours}) & 3.5M & 0.4625   \\
		\bottomrule
	\end{tabular}
	\vspace{-2ex}
\end{table}

Table~\ref{lbl-model-size-inference-timing} summarizes the model size in number of parameters, and reports the inference time for processing the 400 sentences in MAMS-ACSA test dataset. Among non-Bert models, our model has relatively small size. The slightly longer inference time is due to aspect-specific EDU representation computation. A hard sentence often contains multiple EDUs. Compared to Bert-baseline, our model is much smaller and only takes about half of its inference time. The Bert variant of our model has a  longer inference time as our model needs to encode an EDU at a time using Bert, instead of encoding a full sentence at a time as in Bert-baseline.  Overall, our \modelname has a small model size, and achieves good performance  with reasonable inference time.  

\section{Conclusion}
We observe that text span in an EDU tends to express a single aspect and unitary sentiment towards the aspect. Hence, we propose a simple \modelname model to learn aspect-specific representations of EDUs in a sentence. Based on our observation, we apply sparse attention at both word-level and EDU-level, to highlight sentiment-bearing words, and  to constraint an EDU to one aspect. Our model shows improvement over strong baselines on three benchmark datasets. The detailed ablation study also shows that the model behaves as expected. Interestingly, the prediction of aspect labels of EDUs, based on EDU-level attention scores, can be beneficial to other applications like review summarization. 
\bibliographystyle{splncs04}
\bibliography{Sentiment}
\end{document}